\documentclass{article}
\usepackage{ijcai22}

\usepackage{times}
\usepackage{soul}
\usepackage{url}
\usepackage[hidelinks]{hyperref}
\usepackage[utf8]{inputenc}
\usepackage[small]{caption}
\usepackage{graphicx}
\usepackage{amsmath}
\usepackage{amsfonts}
\usepackage{amsthm}
\usepackage{booktabs}
\usepackage{algorithm}
\usepackage{algorithmic}
\usepackage{multirow}
\usepackage{bm}
\usepackage{bbm}
\usepackage{subfigure}
\usepackage{color}
\title{SELC: Self-Ensemble Label Correction Improves Learning with Noisy Labels}

\author{
Yangdi Lu
\and
Wenbo He
\affiliations
McMaster University, Hamilton, Canada \\
\emails
\{luy100, hew11\}@mcmaster.ca
}

\begin{document}

\maketitle

\begin{abstract}
	Deep neural networks are prone to overfitting noisy labels, resulting in poor generalization performance. To overcome this problem, we present a simple and effective method \emph{self-ensemble label correction} (SELC) to progressively correct noisy labels and refine the model. We look deeper into the memorization behavior in training with noisy labels and observe that the network outputs are reliable in the early stage. To retain this reliable knowledge, SELC uses ensemble predictions formed by an exponential moving average of network outputs to update the original noisy labels. We show that training with SELC refines the model by gradually reducing supervision from noisy labels and increasing supervision from ensemble predictions. Despite its simplicity, compared with many state-of-the-art methods, SELC obtains more promising and stable results in the presence of class-conditional, instance-dependent, and real-world label noise. The code is available at {\color{magenta}\href{https://github.com/MacLLL/SELC}{https://github.com/MacLLL/SELC}}.
\end{abstract}
\section{Introduction}
The recent success of deep neural networks (DNNs) for vision tasks owes much to the availability of large-scale, correctly annotated datasets. However, obtaining such high-quality datasets can be extremely expensive, and sometimes even impossible. The common approaches, such as web queries \cite{li2017webvision} and crowdsourcing \cite{song2019selfie}, can easily provide extensive labeled data, but unavoidably introduce \emph{noisy labels}. Existing studies \cite{arpit2017closer,zhang2016understanding} have demonstrated that DNNs can easily overfit noisy labels, which deteriorates the generalization performance. Thus, it is essential to develop noise-robust algorithms for learning with noisy labels.
 
Given a noisy training set consisting of clean samples and mislabeled samples, a common category of approaches \cite{reed2014training,arazo2019unsupervised,zhang2020learning} to mitigating the negative influence of noisy labels is to identify and correct the mislabeled samples. However, the correction procedure in these methods only updates the noisy labels using the model prediction from the most recent training epoch directly, thus it may suffer from the false correction as the model predictions for noisy samples tend to fluctuate. Take a bird image mislabeled as an airplane as an example. During the training, the clean bird samples would encourage the model to predict a given bird image as a bird, while the bird images with airplane labels regularly pull the model back to predict the bird as an airplane. Hence, the model prediction gathered in one training epoch may change back and forth between bird and airplane, resulting in false correction.

We investigate the reason for performance degradation by analyzing the memorization behavior of the DNNs models. We observe that there exists a \emph{turning point} during training. Before the turning point, the model only learns from easy (clean) samples, and thus model prediction is likely to be consistent with clean samples. After the turning point, the model increasingly memorizes hard (mislabeled) samples. Hence model prediction oscillates strongly on clean samples. Triggered by this observation, we seek to make the model retain the early-learning memory for consistent predictions on clean samples even after the turning point.

In this paper, we propose self-ensemble label correction (SELC), which potentially corrects noisy labels during training thus preventing the model from being affected by the noisy labels. SELC leverages the knowledge provided in the model predictions over historical training epochs to form a consensus of prediction (ensemble prediction) before the turning point. We demonstrate that combining ensemble prediction with the original noisy label leads to a better target. Accordingly, the model is gradually refined as the targets become less noisy, resulting in improving performance. However, it is challenging to find the turning point. Existing works estimate the turning point based on a test set or noise information, which are unobservable in practice. We propose a metric to estimate the turning point only using training data, allowing us to select a suitable initial epoch to perform SELC. Overall, our contributions are summarized as follows:

\begin{itemize}
	\item We propose a simple and effective label correction method SELC based on self-ensembling.
	\item We design an effective metric based on unsupervised loss modeling to detect the turning point without requiring the test set and noise information.
	\item SELC achieves superior results and can be integrated with other techniques such as mixup \cite{zhang2018mixup} to further enhance the performance. 
\end{itemize}

\begin{figure*}
	\centering
	\subfigure[Training accuracy]{
		\includegraphics[width=0.32\columnwidth]{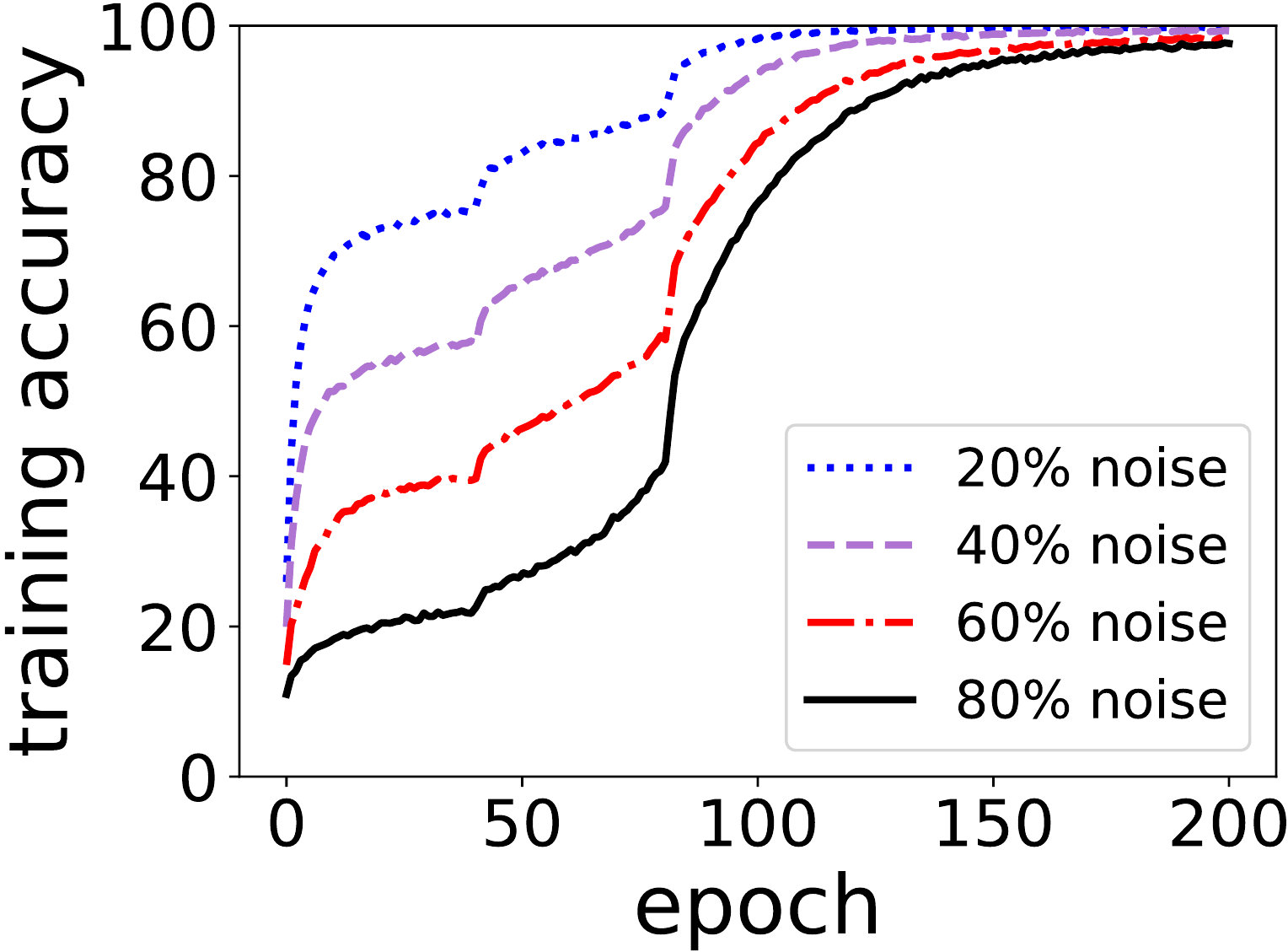}}
	\subfigure[Test accuracy]{
		\includegraphics[width=0.32\columnwidth]{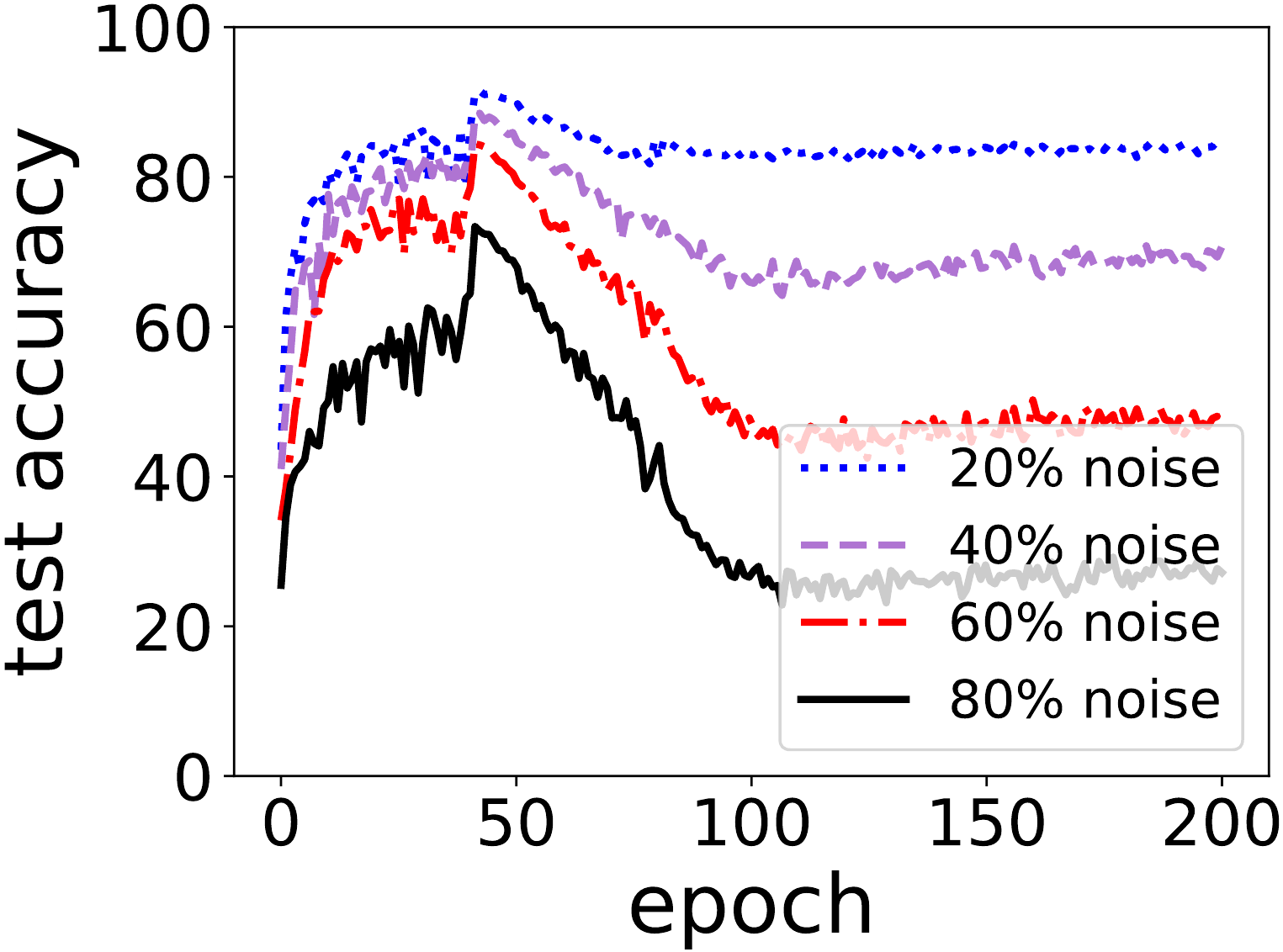}}
	\subfigure[CE]{
		\includegraphics[width=0.32\columnwidth]{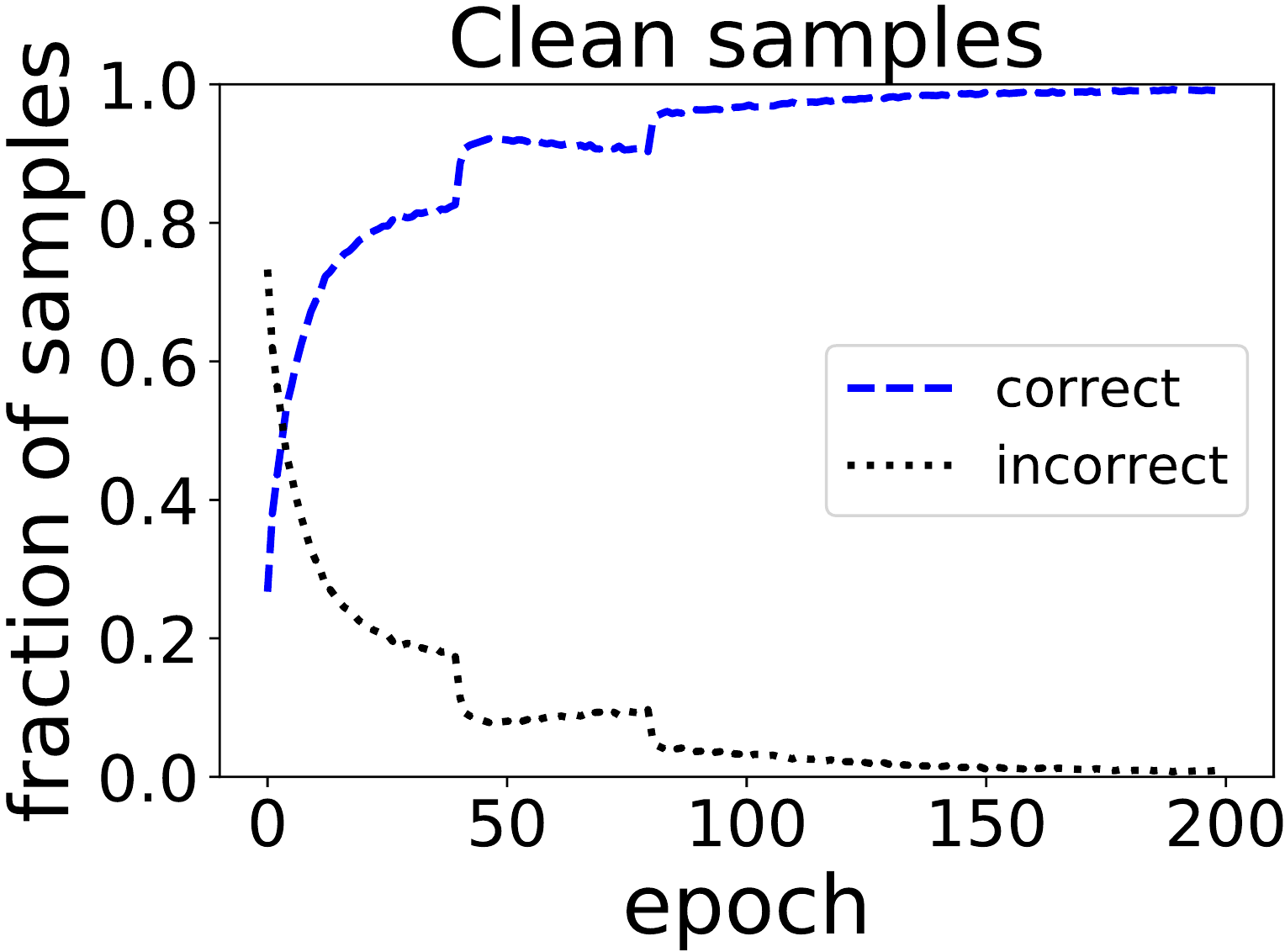}}
	\subfigure[CE]{
		\includegraphics[width=0.32\columnwidth]{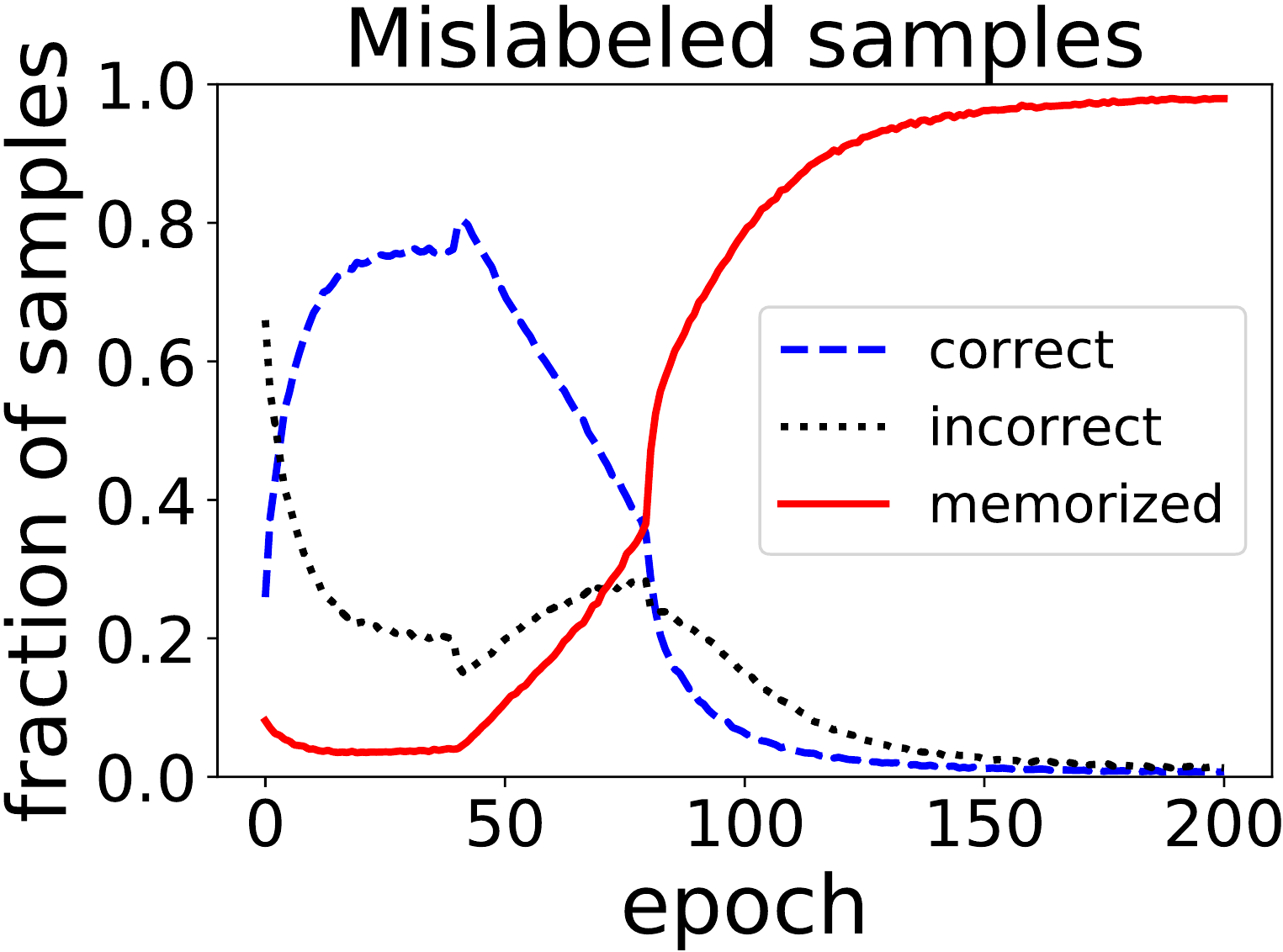}}
	\subfigure[SELC]{
		\includegraphics[width=0.32\columnwidth]{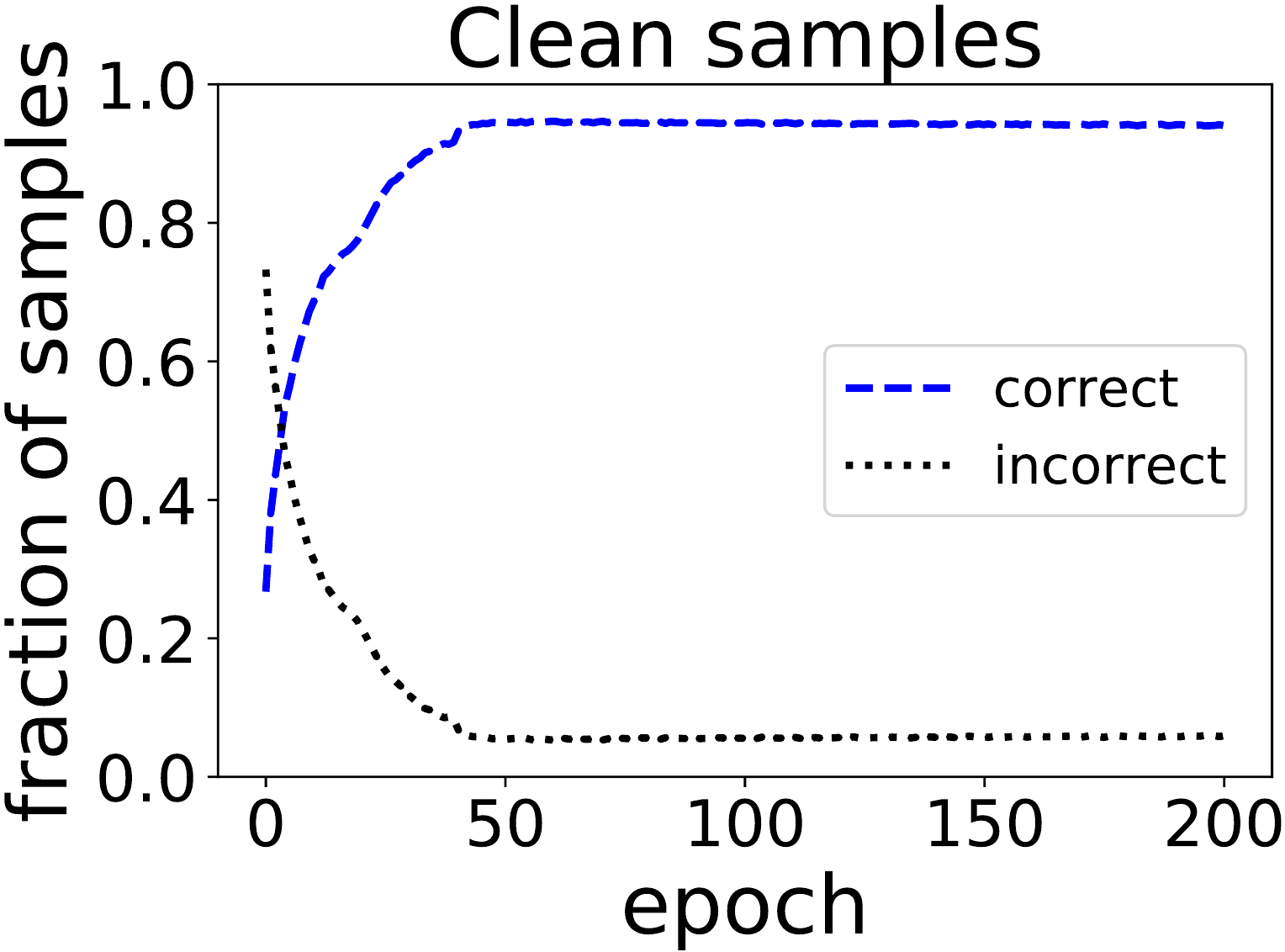}}
	\subfigure[SELC]{
		\includegraphics[width=0.32\columnwidth]{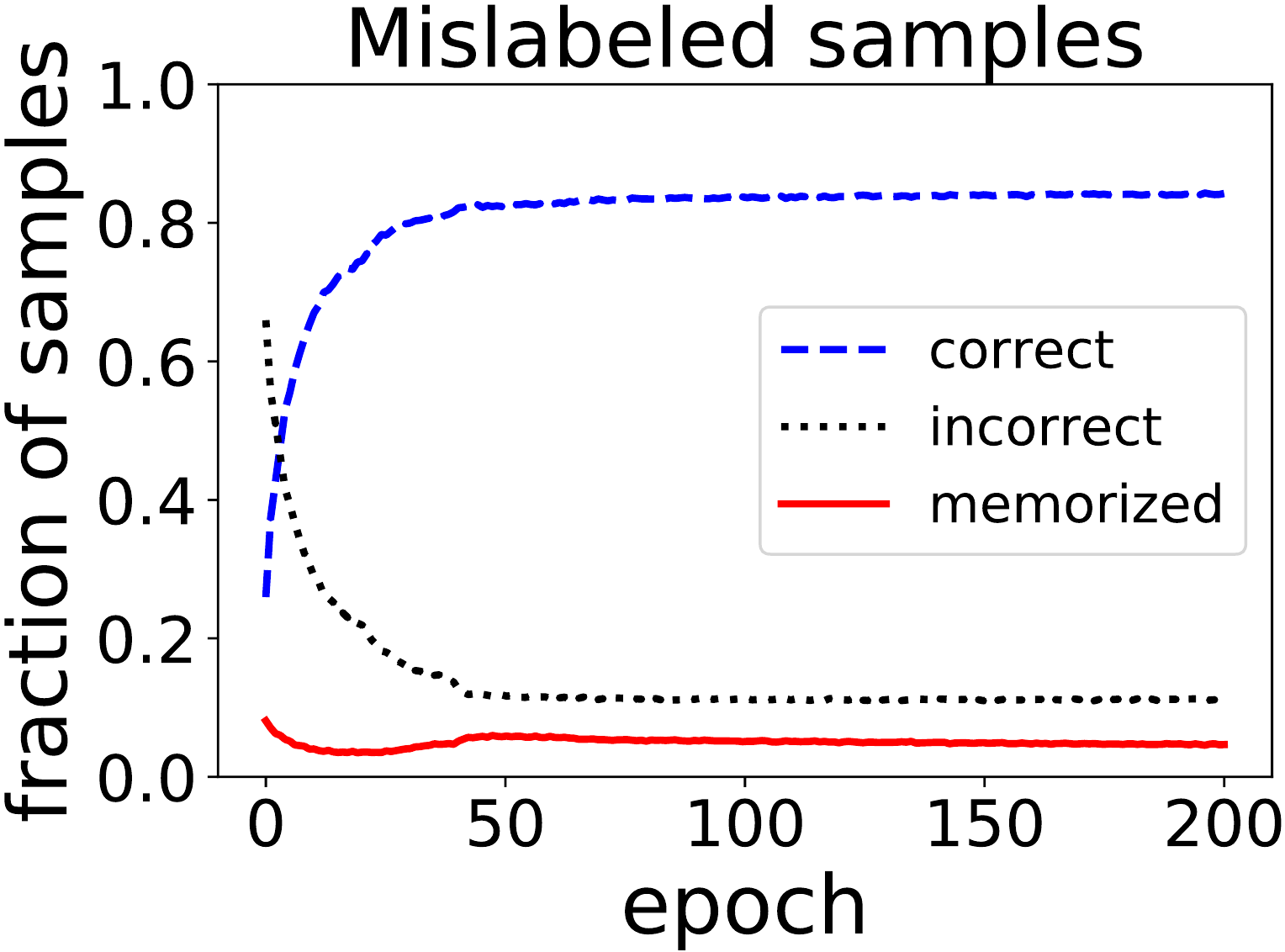}}
	\caption{Plots (a) and (b) show the training and test accuracy on CIFAR-10 with different ratios of label noise using cross-entropy (CE) loss. We investigate the memorization behavior of DNNs on CIFAR-10 with 60\% label noise using CE loss and SELC. Plot (c) and (e) show the fraction of clean samples that are predicted correctly (blue) and incorrectly (black). Plot (d) and (f) show the fraction of mislabeled samples that are predicted correctly (blue), \emph{memorized} (i.e. the prediction equals to the wrong label, shown in red), and incorrectly predicted as neither the true nor the given wrong label (black). Compared to CE, SELC effectively prevents memorization of mislabeled samples and refines the model to attain correct predictions on both clean and mislabeled samples.}
	\label{fig:pre}
\end{figure*}

\section{Related Work}
We briefly discuss the existing noise-robust methods that do not require a small set of clean training data (as
opposed to \cite{xiao2015learning}).

\noindent\textbf{Robust loss functions.} Some methods aim to develop loss functions that are robust to label noise, including GCE \cite{zhang2018generalized}, $\mathcal{L}_\text{DMI}$ \cite{xu2019l_dmi}, SCE \cite{wang2019symmetric} and NCE \cite{ma2020normalized}. \textbf{Loss correction.} \cite{patrini2017making} focus on correcting the loss function explicitly by estimating the noise transition matrix. \noindent\textbf{Sample selection.} Co-training-based methods \cite{han2018co,lu2022ensemble} maintain two networks, and each network is trained on low-risk samples which are selected by its peer network based on the small-loss criterion. \textbf{Regularization.} These methods \cite{liu2020early,lu2021confidence} prevent memorization of mislabeled samples by using a regularizer. \textbf{Label filtering.} SELF \cite{nguyen2020self} filters the mislabeled samples by ensemble predictions to improve the performance. \textbf{Label correction.} Joint Opt \cite{tanaka2018joint} and PENCIL \cite{yi2019probabilistic} replace the noisy labels with soft (i.e. model probability) or hard (i.e. to one-hot vector) pseudo-labels. Bootstrap \cite{reed2014training} and M-correction \cite{arazo2019unsupervised} correct the labels by using a convex combination of noisy labels and the model prediction. PLC \cite{zhang2020learning} updates the noisy labels of high confident samples with model predictions.

Our method is related to label correction. Compared with existing methods, we focus on using ensemble prediction based on historical model outputs to correct the noisy labels, rather than only using prediction from the most recent training epoch. Our approach is straightforward and yields superior performance. Furthermore, our technique can be employed as an add-on component to further enhance the other approaches in challenging cases.

\section{Preliminaries}
\label{sec:pre}
\textbf{Supervised Classification.} Considering a supervised classification problem with $C$ classes, suppose $\mathcal{X}\in\mathbb{R}^{d}$ be the input space, $\mathcal{Y}\in\{0,1\}^{C}$ is the ground-truth label space in an one-hot manner. In practice, the joint distribution $\mathcal{P}$ over $\mathcal{X}\times\mathcal{Y}$ is unknown. We have a training set $D=\{(\bm{x}_{i},\bm{y}_{i})\}^{N}_{i=1}$ which are independently sampled from $\mathcal{P}$. Assume a mapping function class $\mathcal{F}$ wherein each $f:\mathcal{X}\rightarrow\mathbb{R}^{C}$ maps the input space to $C$-dimensional score space, we seek $f^{*}\in\mathcal{F}$ that minimizes an empirical risk $\frac{1}{N}\sum_{i=1}^{N}\ell(\bm{y}_{i},f(\bm{x}_{i}))$ for a certain loss function $\ell$.

\noindent\textbf{Learning with Noisy Labels.} Our goal is to learn from a noisy training distribution $\mathcal{P}_{\eta}$ where the labels are corrupted, with probability $\eta$, from their true distribution $\mathcal{P}$. Given a noisy training set $\hat{D}=\{(\bm{x}_{i},\bm{\hat{y}}_{i})\}^{N}_{i=1}$, the observable noisy label $\bm{\hat{y}}_{i}$ has a probability of $\eta$ to be incorrect. Suppose the mapping function $f$ is a deep neural network classifier parameterized by $\Theta$. $f$ maps an input $\bm{x}_{i}$ to $C$-dimensional logits $\bm{z}_{i}=f(\bm{x}_{i},\Theta)$. We obtain conditional probability of each class by using a softmax function $\mathcal{S}(\cdot)$, thus $\bm{p}_{i}=\mathcal{S}(\bm{z}_{i})$. Then the empirical risk on $\hat{D}$ using cross-entropy loss is
\begin{align}
	\label{eq:ce}
	\mathcal{L}_{\text{ce}}=\frac{1}{N}\sum_{i=1}^{N}\ell_{\text{ce}}(\bm{\hat{y}}_{i},\bm{p_{i}})=-\frac{1}{N}\sum_{i=1}^{N}(\bm{\hat{y}}_{i})^{\top}\log(\bm{p}_{i}).
\end{align}
When optimizing $\mathcal{L}_{\text{ce}}$ by stochastic gradient descent (SGD), the DNNs have been observed to completely fit the training set including mislabeled samples eventually (see Figure \ref{fig:pre} (a)), resulting in the test performance degradation in the later stage of training (see Figure \ref{fig:pre} (b)).

\noindent\textbf{Noise Models.} The generation of real-world label noise is unpredictable, a common methodology to cope with noisy labels is to posit a noise model and design robust algorithms under this model. Then we evaluate the algorithms on the real-world datasets to see their effectiveness. A common noise model is class-conditional noise \cite{natarajan2013learning}, wherein label noise is independent of input features and true label is corrupted by either a \emph{symmetric} or \emph{asymmetric} noise transition matrix (details are in Section \ref{sec:ccn}). Recently, another label noise model, named instance-dependent noise \cite{zhang2020learning,chen2021beyond}, is proposed, in which the noise not only depends on the class but also the input feature.

\section{Our Method}
\label{sec:method}
\subsection{Memorization Behavior}
Our motivation stems from the memorization behavior of DNNs when trained with noisy labels. In Figure \ref{fig:pre} (c), we observe that for clean samples, the model predicts them correctly with the increase of epochs. For mislabeled samples in Figure \ref{fig:pre} (d), the model predicts the true labels correctly for most mislabeled samples in the early stage (high blue line), even though the model begins making incorrect predictions because of the memorization of wrong labels (increasing red line). Since the model predictions are relatively correct for both mislabeled and clean samples in the early stage, can these reliable model predictions help correct the noisy labels?

\subsection{Ensemble Prediction}
To alleviate the impact of noisy labels, existing work Bootstrap \cite{reed2014training} proposes to generate soft target by interpolating between the original noisy distributions and model predictions by $\beta\bm{\hat{y}}+(1-\beta)\bm{p}$, where $\beta$ weights the degree of interpolation. Thus the cross-entropy loss using Bootstrap becomes
\begin{align}
	\label{eq:bs}
	\mathcal{L}_{\text{bs}}=-\frac{1}{N}\sum_{i=1}^{N}\Big(\beta\bm{\hat{y}}_{i}+(1-\beta)\bm{p}_{i}\Big)^{\top}\log(\bm{p}_{i}).
\end{align}
However, applying a static weight (e.g. $\beta=0.8$) to the prediction limits the correction of a hypothetical noisy label. Although another work M-correction \cite{arazo2019unsupervised} makes $\beta$ dynamic for different samples, the one-step correction based solely on the model predictions at the most recent training epoch still easily incurs false correction.	

Since the predictions gathered in a single training epoch for correction is sub-optimal, we generate the ensemble prediction $\tilde{\bm{p}}$ for each sample, aggregating the predictions over multiple previous epochs by exponential moving average. Let's denote the model prediction in epoch $k$ as $\bm{p}_{[k-1]}$. In epoch $k$, we have ensemble prediction 
\begin{align}
	\label{eq:enp}
	\tilde{\bm{p}}_{[k]}=\left\{ \begin{array}{ll}
		\bm{0} & \textrm{if } k =0\\
		\alpha \tilde{\bm{p}}_{[k-1]} + (1-\alpha) \bm{p}_{[k]}, & \textrm{if $k > 0$}  \\
	\end{array} \right. 
\end{align}
where $0\le\alpha<1$ is the momentum. Based on the Eq. (\ref{eq:enp}), we can derive the ensemble prediction in $k$-th epoch as $\tilde{\bm{p}}_{[k]}=\sum_{j=1}^{k}(1-\alpha)\alpha^{k-j}\bm{p}_{[j]}$. Although ensemble prediction requires a new hyperparameter $\alpha$ and auxiliary memory to record, it maintains a more stable and accurate prediction, especially for mislabeled samples. 

\subsection{Self-Ensemble Label Correction}
\label{sec:selc}

\begin{figure*}[t]
	\begin{center}
		\includegraphics[width=1.0\linewidth]{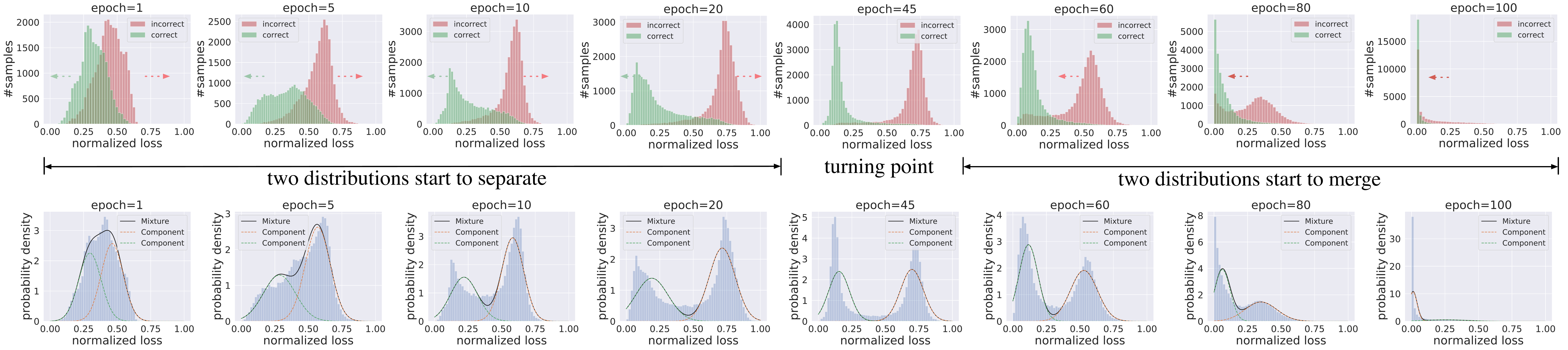}
	\end{center}
	\caption{We train ResNet34 on the CIFAR-10 with 60\% symmetric label noise using cross-entropy loss and investigate the loss distribution. Top row: The normalized loss distribution over different training epochs. Bottom row: The corresponding mixture model and two components after fitting a two-component GMM to loss distribution. }
	\label{fig:tp}
\end{figure*}
We seek to utilize the ensemble predictions to progressively enhance the targets in loss function. There are two options to be considered.
\begin{itemize}
	\item \emph{Option I}. Directly use ensemble prediction as the target.
	\item \emph{Option II}. Preserve the original noisy label, and combine it with ensemble prediction as the target.
\end{itemize}
The first option is widely adopted in semi-supervised learning, as the ensemble prediction learned from the labeled inputs can be used as targets for the unlabeled inputs. However, in the noisy labels setting, the model needs supervisions from noisy labels as no extra clean samples are provided. We would compare these two options in Section \ref{sec:analysis}. In SELC, we choose the second option. Specifically, for each training sample, we initialize the soft target $\bm{t}$ using original noisy label $\bm{\hat{y}}$. Then we update $\bm{t}$ in each training epoch $k$ by
\begin{align}
	\label{eq:lc}
	\bm{t}_{[k]}=\left\{ \begin{array}{ll}
		\bm{\hat{y}} & \textrm{if } k =0\\
		\alpha \bm{t}_{[k-1]} + (1-\alpha) \bm{p}_{[k]}. & \textrm{if $k > 0$}  \\
	\end{array} \right. 
\end{align}
Based on the Eq. (\ref{eq:lc}), we rewrite above equation as 
\begin{align}
	\label{eq:lc_new}
	\bm{t}_{[k]}=\underbrace{\alpha^{k}\bm{\hat{y}}}_\text{first term}+\underbrace{\sum_{j=1}^{k}(1-\alpha)\alpha^{k-j}\bm{p}_{[j]}.}_\text{second term}
\end{align}
The first term preserves the original noisy labels with exponential decaying weights $\alpha^{k}$. The second term is exactly the ensemble prediction at epoch $k$. Therefore, in training epoch $k$, the loss of SELC becomes
\begin{align}
	\label{eq:mtloss}
	\mathcal{L}_{\text{selc}}&=-\frac{1}{N}\sum_{i=1}^{N}\Big(\alpha^{k}\bm{\hat{y}_{i}}+\sum_{j=1}^{k}(1-\alpha)\alpha^{k-j}\bm{p_{i[j]}}\Big)^{\top}\log(\bm{p}_{i}) \nonumber \\
	&= -\frac{1}{N}\sum_{i=1}^{N}\alpha^{k}(\bm{\hat{y}_{i}})^{\top}\log(\bm{p}_{i}) \nonumber \\&\quad\quad
	- \frac{1}{N}\sum_{i=1}^{N}\Big(\sum_{j=1}^{k}(1-\alpha)\alpha^{k-j}\bm{p_{i[j]}}\Big)^{\top}\log(\bm{p}_{i})
\end{align}
where $\bm{p_{i[j]}}$ denotes the model prediction in epoch $j$ for input $\bm{x}_{i}$. The first loss term is actually the cross-entropy loss $\mathcal{L}_\text{ce}$ but weighed by $\alpha^{k}$. With the increase of training epoch $k$, $\alpha^{k}$ becomes smaller. Thus $\mathcal{L}_\text{selc}$ is less and less reliant on original noisy labels. The second loss term maintains an exponential moving average of historical prediction as target, and penalizes model predictions that are inconsistent with this target. As a consequence, SELC effectively prevents memorization of mislabeled samples (low red line in Figure \ref{fig:pre} (f)) and attains superior performance.

\subsection{Estimation of Turning Point}
In semi-supervised learning, the initial epoch to perform ensemble prediction is not crucial as the supervision from the clean set guides the model to predict consistent prediction throughout the training. Comparatively, in our scenario, the model would overfit to noisy labels, causing the model predictions to deteriorate. Therefore, it is essential to select the initial epoch in SELC before the turning point $T$, at which the model starts to memorize mislabeled samples.

We can clearly observe the occurrence of turning point by monitoring the test accuracy drop (Figure \ref{fig:pre} (b)). However, the test set is unobservable in practice. The way to accurately identify the turning point without a test set and noise information remains challenging and underexplored.

In this paper, we propose three metrics and choose the optimal one to estimate the turning point by modeling training samples' loss distribution without requiring a clean test set. Due to the memorization behavior of DNNs, the clean samples tend to have smaller loss values than the mislabeled samples in early stage. We analyze the normalized loss distribution over different training epochs in Figure \ref{fig:tp} top row. Intriguingly, the two distributions are merged at the initialization, then start to separate, but resume merging after the turning point. Therefore, we propose to estimate the turning point by finding the epoch that has the largest distance between two distributions. To model these two distributions, we use two unsupervised learning approaches: Gaussian Mixture Model (GMM) \cite{permuter2006study} and K-Means.

\noindent\textbf{Metric 1 and Metric 2.} We fit a two-component GMM to loss distribution (in Figure \ref{fig:tp} bottom row). The probability density function (pdf) of GMM with $M$ components on the per sample loss value $\ell$ can be defined as $P(\ell) = \sum_{m=1}^{M}\pi_{m}\mathcal{N}(\ell\mid \mu_{m}, \sigma^{2}_{m}), \quad \sum_{m=1}^{M}\pi_{m}=1$, where $\pi_{m}$ is the coefficient for the linear convex combination of each individual pdf $\mathcal{N}(\ell\mid \mu_{m}, \sigma^{2}_{m})$. We use the Expectation-Maximization (EM) algorithm to estimate the $\pi_{m}$, $\mu_{m}$ and $\sigma_{m}^{2}$. For Metric 1, we directly calculate the distance between two components by
\begin{align}
	\mathcal{M}_\text{1}=\mid \mu_{1} - \mu_{2}\mid.
\end{align}
For Metric 2, we calculate the Kullback–Leibler (KL) divergence of two components as distance.
\begin{align}
	\mathcal{M}_\text{2}=\log\frac{\sigma_{2}}{\sigma_{1}}+\frac{\sigma_{1}^{2}+(\mu_{1}-\mu_{2})^{2}}{2\sigma_{2}^{2}}-\frac{1}{2}.
\end{align}

\begin{figure}[t]
	\begin{center}
		\includegraphics[width=1\linewidth]{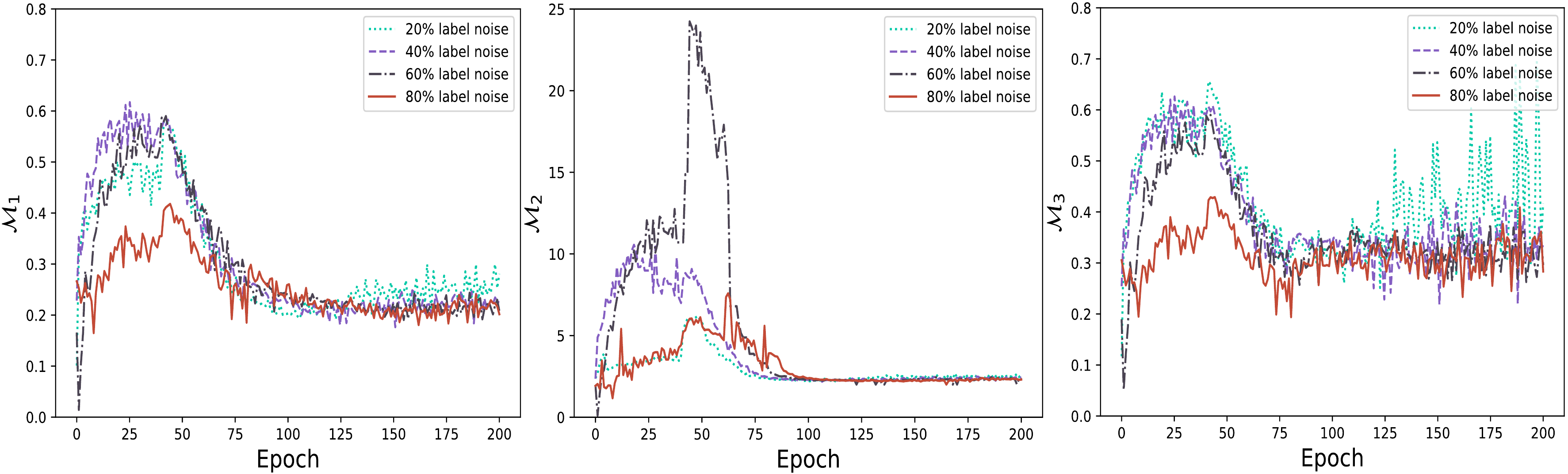}
	\end{center}
	\caption{Three metrics on CIFAR-10 with different ratios of noise.  }
	\label{fig:metric}
\end{figure}

\noindent\textbf{Metric 3.} We fit two clusters by K-Means on the loss distribution. Then we calculate the distance between two cluster centroids $\mathcal{S}_{1}$ and $\mathcal{S}_{2}$ as the Metric 3.
\begin{align}
	\mathcal{M}_\text{3}=\mid\mathcal{S}_{1}-\mathcal{S}_{2} \mid.
\end{align}
When we train the DNNs with noisy labels, we monitor these three metrics. Once they achieve the maximum value, the corresponding epoch is likely to be the turning point $T$. We compare three metrics on CIFAR-10 with label noise in Figure \ref{fig:metric}. $\mathcal{M}_{1}$ is the most reliable and stable one since its corresponding epoch of maximum value precisely aligns with the epoch when test accuracy starts to drop in Figure \ref{fig:pre} (b) in all noise cases. We put pseudocode of SELC in Algorithm \ref{alg:algorithm}. 

\begin{table*}
	
	\centering
	\resizebox{1.0\textwidth}{!}{
		\centering
		\begin{tabular}{ p{40mm} c c c  c c cc c c c c } 
			
			\toprule
			\multicolumn{2}{c}{\multirow{1}{*}{Dataset}} & \multicolumn{5}{c}{CIFAR-10} &\multicolumn{5}{c}{CIFAR-100}\\ \cmidrule(lr){3-7} \cmidrule(lr){8-12}
			\multicolumn{2}{c}{\multirow{1}{*}{Class-conditional noise type}} & \multicolumn{4}{c}{symm} &\multicolumn{1}{c}{asymm} & \multicolumn{4}{c}{symm}&\multicolumn{1}{c}{asymm} \\   \cmidrule(lr){3-6} \cmidrule(lr){7-7} \cmidrule(lr){8-11} \cmidrule(lr){12-12}
			\multicolumn{2}{c}{\multirow{1}{*}{Method/Noise ratio}}& 20\% & \multicolumn{1}{c}{40\%}&\multicolumn{1}{c}{60\%}&\multicolumn{1}{c}{80\%}& \multicolumn{1}{c}{40\%} & \multicolumn{1}{c}{20\%}&\multicolumn{1}{c}{40\%}&\multicolumn{1}{c}{60\%}&\multicolumn{1}{c}{80\%}&\multicolumn{1}{c}{40\%}\\
			\midrule
			\multirow{1}{*}{Cross Entropy} &  & 86.98 $\pm$ 0.12 &81.88 $\pm$ 0.29&74.14 $\pm$ 0.56&53.82 $\pm$ 1.04&80.11 $\pm$ 1.44&58.72 $\pm$ 0.26&48.20 $\pm$ 0.65&37.41 $\pm$ 0.94&18.10 $\pm$ 0.82&42.74 $\pm$ 0.61\\ 
			\multirow{1}{*}{Bootstrap \cite{reed2014training}} && 86.23 $\pm$ 0.23 &82.23 $\pm$ 0.37&75.12 $\pm$ 0.56&54.12 $\pm$ 1.32&81.21 $\pm$ 1.47&58.27 $\pm$ 0.21&47.66 $\pm$ 0.55&34.68 $\pm$ 1.10&21.64 $\pm$ 0.97&45.12 $\pm$ 0.57\\ 
			
			\multirow{1}{*}{Forward \cite{patrini2017making}} && 87.99 $\pm$ 0.36 &83.25 $\pm$ 0.38&74.96 $\pm$ 0.65&54.64 $\pm$ 0.44&83.55 $\pm$ 0.58&39.19 $\pm$ 2.61&31.05 $\pm$ 1.44&19.12 $\pm$ 1.95&8.99 $\pm$ 0.58&34.44 $\pm$ 1.93\\

			\multirow{1}{*}{GCE \cite{zhang2018generalized}} & &89.83 $\pm$ 0.20 &87.13 $\pm$ 0.22&82.54 $\pm$ 0.23&64.07 $\pm$ 1.38&76.74 $\pm$ 0.61&66.81 $\pm$ 0.42&61.77 $\pm$ 0.24&53.16 $\pm$ 0.78&29.16 $\pm$ 0.74&47.22 $\pm$ 1.15\\
			
			\multirow{1}{*}{Mixup \cite{zhang2018mixup}}& &93.58&89.46&78.32 &66.32& 81.66 & 69.31& 58.12&41.10&18.77&49.61\\
			
			\multirow{1}{*}{Joint Opt \cite{tanaka2018joint}} & & 92.25 &90.79&86.87&69.16&-&58.15&54.81&47.94&17.18&-\\
			\multirow{1}{*}{PENCIL \cite{yi2019probabilistic} } & & - &-&-&-&91.01&-&69.12 $\pm$ 0.62&57.70 $\pm$ 3.86&fail&63.61 $\pm$ 0.23\\
			
			\multirow{1}{*}{NLNL \cite{kim2019nlnl}} & & 94.23 &92.43&88.32&-&89.86&71.52&66.39&56.51&-&45.70\\
			
			\multirow{1}{*}{SCE \cite{wang2019symmetric}} & & 89.83 $\pm$ 0.20 &87.13 $\pm$ 0.26&82.81 $\pm$ 0.61&68.12 $\pm$ 0.81&82.51 $\pm$ 0.45&70.38 $\pm$ 0.13&62.27 $\pm$ 0.22&54.82 $\pm$ 0.57&25.91 $\pm$ 0.44&69.32 $\pm$ 0.87\\
			
			\multirow{1}{*}{M-correction \cite{arazo2019unsupervised}} & & - &92.30&86.10&74.10&-&-&70.10&59.50&\textbf{39.50}&-\\
			
			\multirow{1}{*}{DAC \cite{thulasidasan2019combating}} & & 92.91 &90.71&86.30&74.84&-&73.55&66.92&57.17&32.16&-\\
			
			\multirow{1}{*}{SELF \cite{nguyen2020self}} & & - &91.13&-&63.59&-&-&66.71&-&35.56&-\\

			\multirow{1}{*}{NCE+RCE \cite{ma2020normalized}} & & - &86.02 $\pm$ 0.09&79.78 $\pm$ 0.50&52.71 $\pm$ 1.90&79.59 $\pm$ 0.40&-&59.48 $\pm$ 0.56&47.12 $\pm$ 0.62&25.80 $\pm$ 1.12&46.69 $\pm$ 0.96\\

			\multirow{1}{*}{ELR \cite{liu2020early}} & & 91.16 $\pm$ 0.08&89.15 $\pm$ 0.17&86.12 $\pm$ 0.49&73.86 $\pm$ 0.61&90.12 $\pm$ 0.47&74.21 $\pm$ 0.22&68.28 $\pm$ 0.31&59.28 $\pm$ 0.67&29.78 $\pm$ 0.56&73.26 $\pm$ 0.64\\
			\midrule
			\multirow{1}{*}{SELC (Ours)}& &93.09 $\pm$ 0.02&91.18 $\pm$ 0.06&87.25 $\pm$ 0.09 &74.13 $\pm$ 0.14& 91.05 $\pm$ 0.11& 73.63 $\pm$ 0.07&68.46 $\pm$ 0.10&59.41 $\pm$ 0.06&32.63 $\pm$ 0.06&70.82 $\pm$ 0.09\\ 
			\multirow{1}{*}{SELC+ (Ours)}& &\textbf{94.97 $\pm$ 0.04}&\textbf{93.12 $\pm$ 0.08}&\textbf{90.46 $\pm$ 0.12} &\textbf{78.62 $\pm$ 0.24}& \textbf{92.92 $\pm$ 0.10}& \textbf{76.39 $\pm$ 0.15}&\textbf{71.73 $\pm$ 0.11}&\textbf{64.49 $\pm$ 0.15}&37.18 $\pm$ 0.50&\textbf{73.58 $\pm$ 0.11}\\

			\bottomrule
			
		\end{tabular}
		
	}
	
	\caption{Test accuracy (\%) on CIFAR-10/100 with various ratios of class-conditional label noise injected to the training set. All methods use the same backbone ResNet34. The average accuracy and standard deviation over 3 trials are reported. The best results are in \textbf{bold}. } 
	\label{tab:ccn}	
\end{table*}

\begin{table}
	\centering
	\resizebox{0.49\textwidth}{!}{
		\begin{tabular}{lcccc}
			
			\toprule
			Method/Noise ratio  & 10\% & 20\% & 30\% & 40\% \\
			\midrule
			Cross Entropy & 91.25 $\pm$ 0.27 &86.34 $\pm$ 0.11 & 80.87 $\pm$ 0.05 & 75.68 $\pm$ 0.29\\
			Forward \cite{patrini2017making} & 91.06 $\pm$ 0.02 &86.35 $\pm$ 0.11 & 78.87 $\pm$ 2.66 & 71.12 $\pm$ 0.47\\
			Co-teaching \cite{han2018co}& 91.22 $\pm$ 0.25 &87.28 $\pm$ 0.20 & 84.33 $\pm$ 0.17 & 78.72 $\pm$ 0.47\\
			GCE \cite{zhang2018generalized}& 90.97 $\pm$ 0.21 &86.44 $\pm$ 0.23 & 81.54 $\pm$ 0.15 & 76.71 $\pm$ 0.39\\
			DAC \cite{thulasidasan2019combating} & 90.94 $\pm$ 0.09 &86.16 $\pm$ 0.13 & 80.88 $\pm$ 0.46 & 74.80 $\pm$ 0.32\\
			DMI \cite{xu2019l_dmi} & 91.26 $\pm$ 0.06 &86.57 $\pm$ 0.16 & 81.98 $\pm$ 0.57 & 77.81 $\pm$ 0.85\\
			SEAL \cite{chen2021beyond}& 91.32 $\pm$ 0.14 &87.79 $\pm$ 0.09 & 85.30 $\pm$ 0.01 & 82.98 $\pm$ 0.05\\
			\midrule
			SELC (Ours) & \textbf{91.63 $\pm$ 0.15 }&\textbf{88.33 $\pm$ 0.16} & \textbf{86.28 $\pm$ 0.22} &\textbf{ 84.23 $\pm$ 0.37}\\
			\bottomrule
		\end{tabular}
		
	}
	\caption{Test accuracy (\%) on CIFAR-10 under instance-dependent label noise from SEAL with different noise ratios. All the compared methods do not use mixup to boost the performance.}
	\label{tab:idn1}
\end{table}

\begin{table*}
	\centering
	\resizebox{1\textwidth}{!}{
		\begin{tabular}{lccccccc|c}
			
			\toprule
			\multirow{2}{*}{Dataset}  & \multirow{2}{*}{Noise} & \multirow{2}{*}{Cross Entropy} & Co-teaching+ & GCE & SCE& LRT & PLC & \multirow{2}{*}{SELC (ours)} \\
			& & & \cite{yu2019does} & \cite{zhang2018generalized} & \cite{wang2019symmetric}& \cite{zheng2020error} & \cite{zhang2020learning} &  \\
			\midrule
			\multirow{6}{*}{CIFAR-10}  & Type-I (35\%) & 78.11 $\pm$ 0.74 & 79.97 $\pm$ 0.15 & 80.65 $\pm$ 0.39 & 79.76 $\pm$ 0.72 & 80.98 $\pm$ 0.80 & 82.80 $\pm$ 0.27& \textbf{86.97 $\pm$ 0.15} \\
			& Type-I (70\%) & 41.98 $\pm$ 1.96 & 40.69 $\pm$ 1.99 & 36.52 $\pm$ 1.62 & 36.29 $\pm$ 0.66& 41.52 $\pm$ 4.53 & 42.74 $\pm$ 2.14 & \textbf{43.78 $\pm$ 2.64} \\
			& Type-II (35\%) & 76.65 $\pm$ 0.57 & 77.34 $\pm$ 0.44 & 77.60 $\pm$ 0.88 & 77.92 $\pm$ 0.89& 80.74 $\pm$ 0.25 & 81.54 $\pm$ 0.47 & \textbf{87.06 $\pm$ 0.20}\\
			& Type-II (70\%) & 45.57 $\pm$ 1.12 & 45.44 $\pm$ 0.64 & 40.30 $\pm$ 1.46 & 41.11 $\pm$ 1.92& 44.67 $\pm$ 3.89 & 46.04 $\pm$ 2.20 & \textbf{46.79 $\pm$ 3.06}\\
			& Type-III (35\%) & 76.89 $\pm$ 0.79& 78.38 $\pm$ 0.67 & 79.18 $\pm$ 0.61 & 78.81 $\pm$ 0.29& 81.08 $\pm$ 0.35 & 81.50 $\pm$ 0.50 & \textbf{87.31 $\pm$ 0.18}\\
			& Type-III (70\%) & 43.32 $\pm$ 1.00 & 41.90 $\pm$ 0.86 & 37.10 $\pm$ 0.59 & 38.49 $\pm$ 1.46& 44.47 $\pm$ 1.23 & 45.05 $\pm$ 1.13 & \textbf{45.57 $\pm$ 1.71}\\
			
			\midrule
			\multirow{6}{*}{CIFAR-100}  & Type-I (35\%) & 57.68 $\pm$ 0.29 & 56.70 $\pm$ 0.71 & 58.37 $\pm$ 0.18 & 55.20 $\pm$ 0.33 & 56.74 $\pm$ 0.34 & 60.01 $\pm$ 0.43& \textbf{65.72 $\pm$ 0.17} \\
			& Type-I (70\%) & 39.32 $\pm$ 0.43 & 39.53 $\pm$ 0.28 & 40.01 $\pm$ 0.71 & 40.02 $\pm$ 0.85& 45.29 $\pm$ 0.43 & 45.92 $\pm$ 0.61 & \textbf{49.72 $\pm$ 0.15}\\
			& Type-II (35\%) & 57.83 $\pm$ 0.25 & 56.57 $\pm$ 0.52 & 58.11 $\pm$ 1.05 & 56.10 $\pm$ 0.73& 57.25 $\pm$ 0.68 & 63.68 $\pm$ 0.29 & \textbf{66.79 $\pm$ 0.18} \\
			& Type-II (70\%) & 39.30 $\pm$ 0.32 & 36.84 $\pm$ 0.39 & 37.75 $\pm$ 0.46 & 38.45 $\pm$ 0.45& 43.71 $\pm$ 0.51 & 45.03 $\pm$ 0.50 & \textbf{52.65 $\pm$ 0.26} \\
			& Type-III (35\%) & 56.07 $\pm$ 0.79 & 55.77 $\pm$ 0.98 & 57.51 $\pm$ 1.16 & 56.04 $\pm$ 0.74& 56.57 $\pm$ 0.30 & 63.68 $\pm$ 0.29 & \textbf{66.41 $\pm$ 0.17} \\
			& Type-III (70\%) & 40.01 $\pm$ 0.18 & 35.37 $\pm$ 2.65 & 40.53 $\pm$ 0.60 & 39.94 $\pm$ 0.84& 44.41 $\pm$ 0.19 & 44.45 $\pm$ 0.62 & \textbf{49.85 $\pm$ 0.36} \\
			\bottomrule
		\end{tabular}
		
	}
	\caption{Test accuracy (\%) on CIFAR under different types of PMD noise with various levels. The average accuracy and standard deviation over 3 trials are reported. All above methods do not use mixup to boost the performance for fair comparison. The best results are in \textbf{bold}. }
	\label{tab:idn2}
\end{table*}

\begin{table}[t]
	\centering
	\resizebox{0.3\textwidth}{!}{
		\begin{tabular}{lc}
			\toprule
			Method  & Accuracy \\
			\midrule
			Cross Entropy        & 79.40 $\pm$ 0.14 \\
			
			Nested \cite{chen2021boosting} & 81.30 $\pm$ 0.60\\
			SELFIE \cite{song2019selfie} & 81.80 $\pm$ 0.09 \\
			PLC \cite{zhang2020learning} & 83.40 $\pm$ 0.43 \\
			\midrule
			SELC (ours)  & \textbf{83.73 $\pm$  0.06}  \\
			\bottomrule
		\end{tabular}
	}
	\caption{The accuracy (\%) results on ANIMAL-10N.}
	\label{tab:animal10n}
\end{table}
\begin{table}[t]
	\centering
	\resizebox{0.3\textwidth}{!}{
		\begin{tabular}{lc}
			\toprule
			Method  & Accuracy \\
			\midrule
			Cross Entropy        & 68.94  \\
			Forward \cite{patrini2017making}      & 69.84      \\
			SEAL \cite{chen2021beyond} & 70.63 \\
			SCE \cite{wang2019symmetric} & 71.02 \\
			LRT \cite{zheng2020error}    & 71.74    \\
			DMI \cite{xu2019l_dmi} & 72.27 \\
			ELR \cite{liu2020early} & 72.87\\
			Nested \cite{chen2021boosting} & 73.10\\
			PENCIL \cite{yi2019probabilistic} & 73.49 \\
			PLC \cite{zhang2020learning} & \textbf{74.02} \\
			\midrule
			SELC (ours)  & 74.01    \\
			\bottomrule
		\end{tabular}
	}
	\caption{The accuracy (\%) results on Clothing1M.}
	\label{tab:clothing1m}
\end{table}

\begin{table}
	\centering
	\resizebox{0.45\textwidth}{!}{
		\begin{tabular}{lcccc}
			\toprule
			\multirow{2}{*}{Method} & \multicolumn{2}{c}{Webvision} &\multicolumn{2}{c}{ILSVRC12}\\
			& top1 & top5 &top1 &top5\\
			\midrule
			Forward \cite{patrini2017making}        & 61.12&82.68 & 57.36&82.36  \\
			Co-teaching \cite{han2018co} & 63.58 & 85.20 & 61.48 & 84.70 \\
			Iterative-CV \cite{chen2019understanding} &65.24 & 85.34 & 61.60 & 84.98\\
			RSL \cite{gui2021towards} &65.64 & 85.72 & 62.04 & 84.84\\
			CRUST \cite{mirzasoleiman2020coresets} & 72.40 & 89.56 & 67.36 & 87.84 \\
			\midrule
			SELC (ours)  & \textbf{74.38} & \textbf{90.66} &\textbf{70.85} &  \textbf{90.74}  \\
			\bottomrule
		\end{tabular}
	}
	\caption{The accuracy (\%) results on (mini) Webvision.} 
	\label{tab:webvision}
\end{table}

\begin{algorithm}[tb]
	\caption{SELC pseudocode.}
	\label{alg:algorithm}
	\textbf{Input}: DNNs $f(\Theta)$, training data $\hat{D}=\{(\bm{x}_{i},\bm{\hat{y}}_{i})\}^{N}_{i=1}$, Estimated turning point $T$, total epoch $T_\text{max}$, hyperparameter $\alpha$ \\
	\textbf{Output}: Optimized DNN $f(\Theta^{*})$
	\begin{algorithmic}[1] 
		\STATE Let $\bm{t}=\bm{\hat{y}}$.
		\STATE Select an initial epoch $T_{e}<T$ (e.g. $T_{e}=T-10$). 
		\WHILE{epoch $e<T_\text{max}$}
		\IF {epoch $e<T_{e}$}
		\STATE Train $f(\Theta)$ by CE loss in Eq. (\ref{eq:ce}) using SGD.
		\ELSE 
		\STATE Update $\bm{t}$ by \eqref{eq:lc_new}.
		\STATE Train $f(\Theta)$ by SELC loss in Eq. (\ref{eq:mtloss}) using SGD.
		\ENDIF
		\ENDWHILE
	\end{algorithmic}
\end{algorithm}

\section{Experiments}
This section, first, investigates the effectiveness of the proposed SELC for classification with class-conditional noise (Section \ref{sec:ccn}), instance-dependent noise (Section \ref{sec:fdn}) and real-world noise (Section \ref{sec:rn}). This is followed by several empirical analyses (Section \ref{sec:analysis}) to shed light on SELC.

\subsection{Class-conditional Label Noise}
\label{sec:ccn}

\textbf{Datasets and Networks.} We conduct the experiments with class-conditional label noise on CIFAR-10 and CIFAR-100 \cite{krizhevsky2009learning}. Given these two datasets are initially clean, we follow \cite{patrini2017making} to inject noise by label transition matrix $\mathbf{Q}$, where $\mathbf{Q}_{ij}=\Pr[\hat{y}=j\mid y=i]$ denotes the probability that noisy label $\hat{y}$ is flipped from true label $y$. We evaluate SELC in two types of noise: \emph{symmetric} and \emph{asymmetric}. Symmetric noise is generated by replacing the labels for a percentage of the training data with all possible labels uniformly. Asymmetric noise is designed to mimic the structure of real-world label noise, where the annotators are more likely to make mistakes only within very similar classes (e.g. \emph{deer} $\rightarrow$ \emph{horse} and \emph{cat} $\leftrightarrow$ \emph{dog}). We use the ResNet34 \cite{he2016deep} as backbone for both datasets, and train the model using SGD with a momentum of 0.9, a weight decay of 0.001, and a batch size of 128. The network is trained for 200 epochs. We set the initial learning rate as 0.02, and reduce it by a factor of 10 after 40 and 80 epochs. We fix hyperparameter $\alpha=0.9$. More discussions on $\alpha$ are in Section \ref{sec:analysis}. Note that we do not perform early stopping since we don’t assume the presence of clean validation data. All test accuracy are recorded from the last epoch of training. 

\noindent\textbf{Improving Other Methods.} In comparison to the original noisy labels, we obtain cleaner targets $\bm{t}$ after using SELC. Therefore, SELC can be easily integrated with other methods. For fair comparison with the existing approaches (e.g. M-correction and NLNL) that use mixup or multiple stages of training to boost the performance, we propose SELC+ which uses the corrected labels from SELC to retrain an initialized DNNs using mixup. 

\noindent\textbf{Results.} Table \ref{tab:ccn} shows the results on CIFAR with different types and levels of class-conditional label noise. SELC achieves excellent performance compared to the methods that only modify the training loss without extra techniques to boost the performance. When integrated with mixup data augmentation, SELC+ achieves the best performance across most noise ratios, demonstrating the effectiveness of the proposed method on class-conditional label noise.

\subsection{Instance-dependent Label Noise}
\label{sec:fdn}
\textbf{Datasets and Networks.} We follow the recent works SEAL \cite{chen2021beyond} and PLC \cite{zhang2020learning} to inject instance-dependent label noise to CIFAR. SEAL generates the controllable label noise based on the assumption that `hard' (low confidence) samples are more likely to be mislabeled. PLC introduces Polynomial Margin Diminishing (PMD) noise which allows arbitrary noise strength in a wide buffer near the decision boundary. For fair comparison with SEAL, we use the same network architecture Wide ResNet28$\times$10. As for PMD noise, we use the same network architecture PreAct ResNet34 as PLC. 

\noindent\textbf{Results.} Table \ref{tab:idn1} shows the results on instance-dependent label noise from SEAL. Our approach consistently achieves the best generalization performance over different noise ratios. The larger the noise ratio is, the more improvement SELC obtains. Table \ref{tab:idn2} lists the performance of different methods under three types of PMD noise at noise level 35\% and 70\%. We observe that the proposed method outperforms baselines across different noise settings. When the noise level is high, performances of a few baselines deteriorate and become worse than the standard (CE) approach. In contrast, the improvement of SELC is substantial ($\sim$10\% in accuracy) for the more challenging CIFAR-100 with 70\% label noise.

\subsection{Real-world Label Noise}
\label{sec:rn}
\textbf{Datasets and Networks.} We use ANIMAL-10N \cite{song2019selfie}, Clothing1M \cite{xiao2015learning} and Webvision \cite{li2017webvision} to evaluate the performance of SELC under the real-world label noise settings. ANIMAL-10N contains human-labeled online images for 10 animals with confusing appearance. The estimated label noise rate is 8\%. Clothing1M consists of 1 million images collected from online shopping websites with labels generated from surrounding texts. The estimated label noise rate is 38.5\%. WebVision contains 2.4 million images crawled from the web using the 1,000 concepts in ImageNet ILSVRC12. The estimated label noise rate is 20\%. For ANIMAL-10N, we use VGG-19 with batch normalization. For Clothing1M, we use ResNet50 pretrained on ImageNet. For Webvision, we use InceptionResNetV2. Note that all the compared method do not use mixup to boost the performance for fair comparison. 

\noindent\textbf{Results.} Table \ref{tab:animal10n}, Table \ref{tab:clothing1m} and Table \ref{tab:webvision}   show the results on ANIMAL-10N, Clothing1M and Webvision respectively. On ANIMAL-10N and Webvision, our approach outperforms the existing baselines. On Clothing1M, SELC achieves the comparable performance to PLC, despite its simplicity.

\begin{figure}[t]
	\begin{center}
		\subfigure[Correction accuracy]{
			\includegraphics[width=0.46\columnwidth]{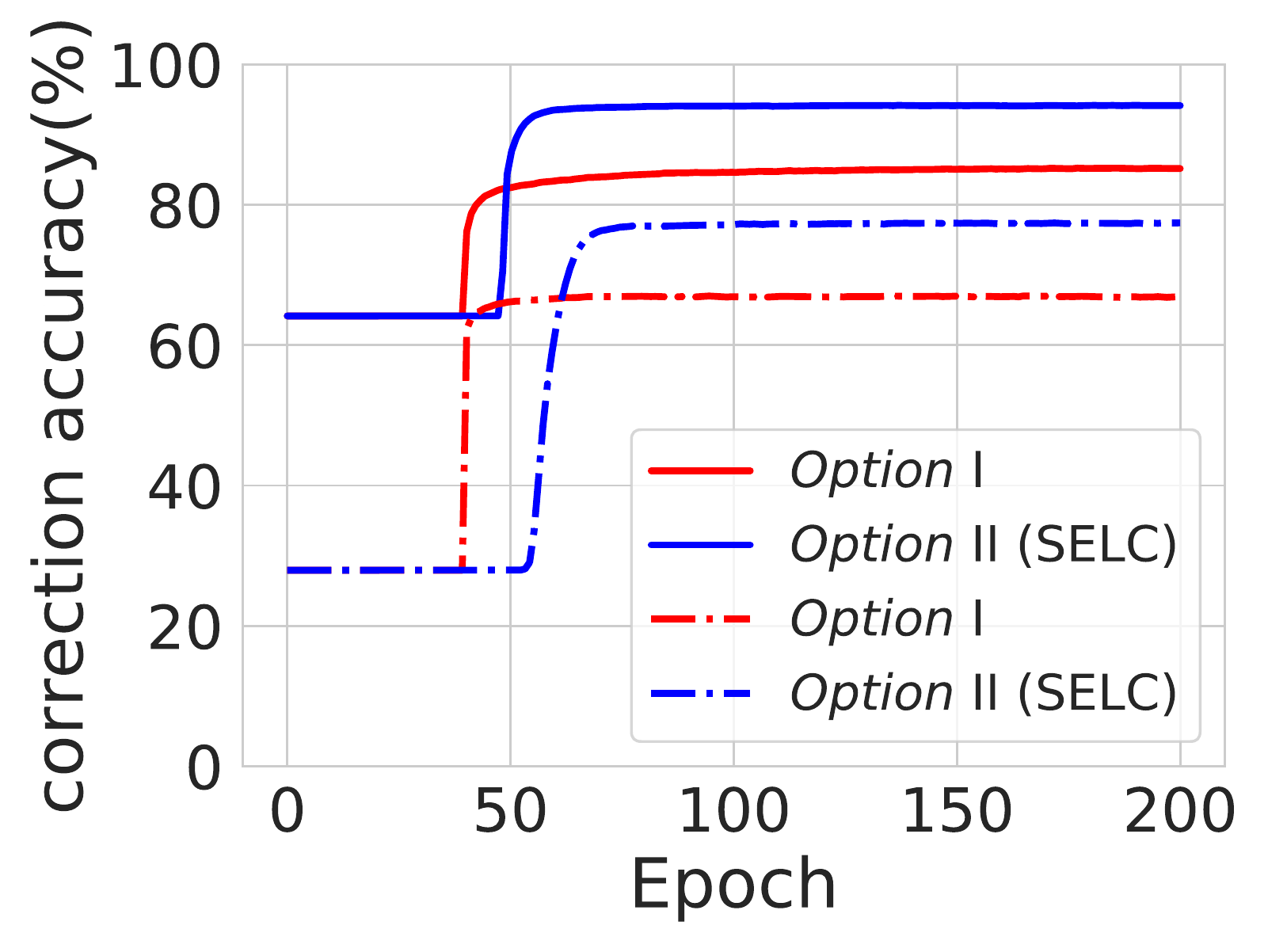}}
		\subfigure[Sensitivity of $\alpha$]{
			\includegraphics[width=0.46\columnwidth]{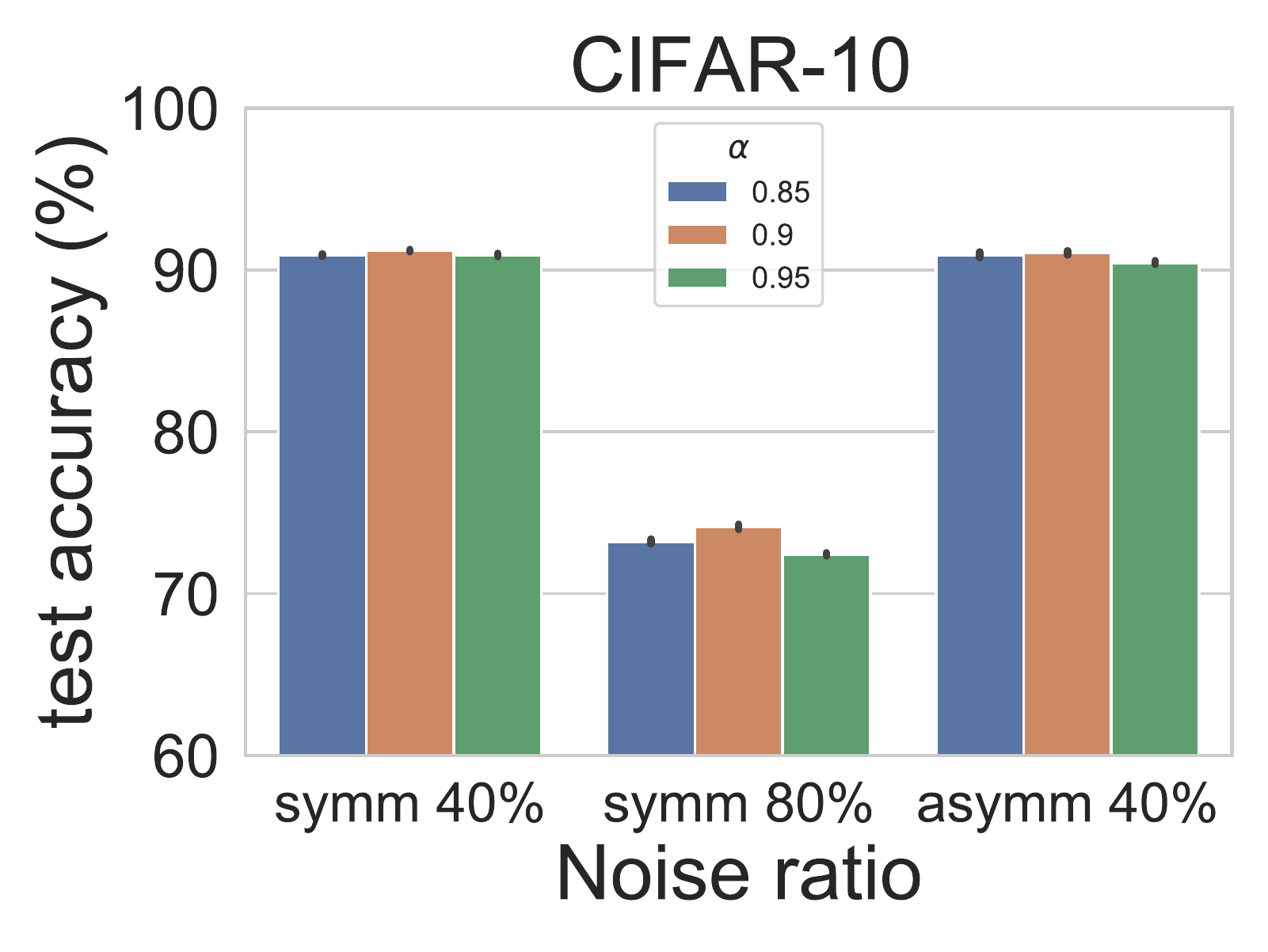}}
	\end{center}
	\vspace{-1em}
	\caption{(a). Correction accuracy of \emph{Option I} and \emph{Option II} on CIFAR-10 with 40\% (solid lines) and 80\% (doted lines) symmetric noise. (b). Sensitivity of $\alpha$.  }
	\label{fig:ca_alpha}
\end{figure}

\begin{figure}[t]
	\begin{center}
		\subfigure[Before using SELC]{
			\includegraphics[width=0.47\columnwidth]{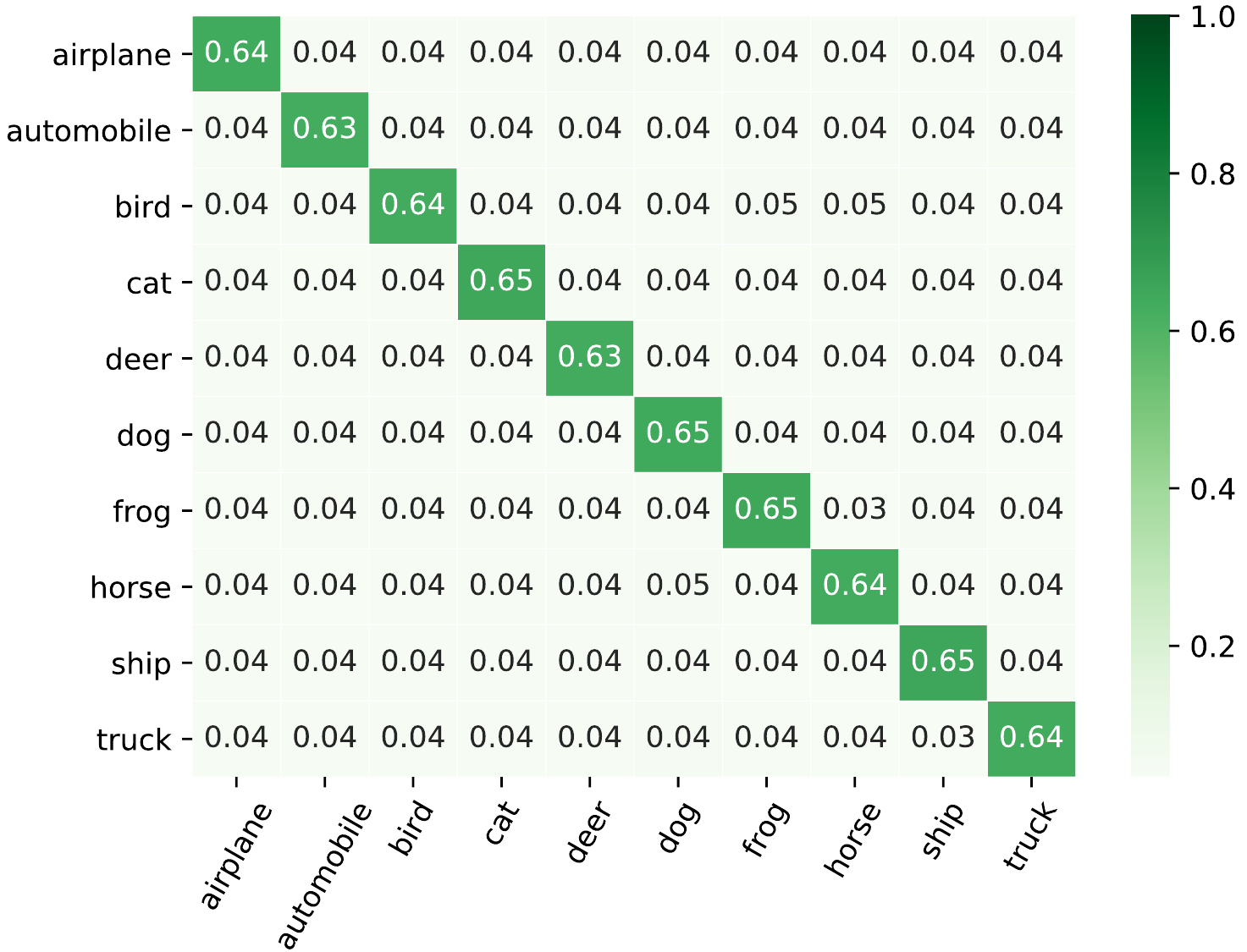}}
		\subfigure[After using SELC]{
			\includegraphics[width=0.47\columnwidth]{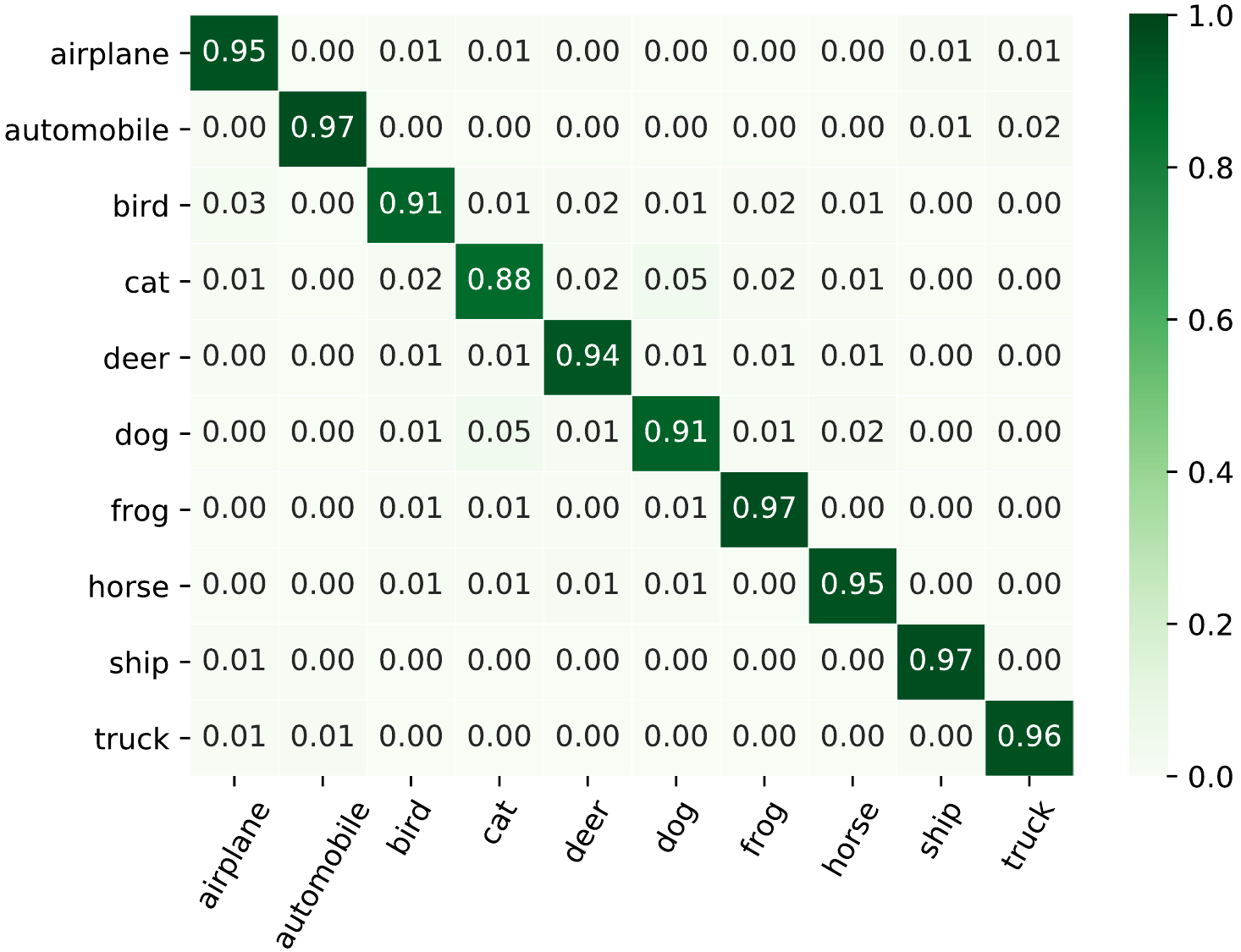}}
	\end{center}
	\vspace{-1em}
	\caption{(a). Confusion matrix of original noisy label w.r.t true labels on CIFAR-10 with 40\% symmetric label noise. (b). Confusion matrix of corrected labels w.r.t true labels after using SELC.}
	\label{fig:confusion}
\end{figure}

\subsection{Empirical Analysis}
\label{sec:analysis}

\noindent\textbf{Correction Accuracy.} The key idea of SELC is to correct the original noisy labels. We analyze the quality of the new target $\bm{t}$ by calculating its correction accuracy: $\frac{1}{N}\sum_{i}^{N}1\{\arg\max \bm{y}_{i}=\arg\max \bm{t}_{i}\}$, where $\bm{y}_{i}$ is the true label of $\bm{x}_{i}$. Figure \ref{fig:ca_alpha} (a) shows the correction accuracy of the Option I and Option II (SELC). We observe SELC achieves higher accuracy than Option I and correction accuracy is stable with the increase of training epochs. Figure \ref{fig:confusion} shows the confusion matrix of corrected labels w.r.t the true labels on CIFAR-10 with 40\% symmetric label noise. SELC corrects the noisy labels impressively well for all classes.

\noindent\textbf{Hyperparameter Sensitivity.} In SELC, we have one hyperparameter $\alpha$ in Eq. (\ref{eq:mtloss}) to control the momentum in self-ensemble. Figure \ref{fig:ca_alpha} (b) shows the test accuracy when using different $\alpha\in\{0.85,0.9,0.95\}$ on CIFAR-10. We observe that the sensitivity to hyperparameter $\alpha$ is quite mild.

\section{Conclusion}
We propose a simple and effective method SELC to improve learning with noisy labels. SELC leverages the model predictions of previous epochs to correct the noisy labels, thus preventing the model from overfitting to noisy labels. By evaluating SELC on different types of label noise, we observe its superior performance over existing approaches.

\small
\bibliographystyle{named}
\bibliography{ijcai22}

\appendix

\section{More Details on Experiments}
We summarize the datasets in Table \ref{table:datasets}.

\begin{table*}
	\begin{center}
		\resizebox{0.8\textwidth}{!}{
			\centering
			\begin{tabular}{c c c c  c c  c} 
				\toprule
				Dataset & \# of train & \# of val & \# of test & \# of classes & input size& Noise rate (\%) \\
				\midrule
				\multicolumn{7}{c}{\multirow{1}{*}{Datasets with clean annotation}} \\
				\midrule
				CIFAR-10 & 50K& - & 10K& 10 & 32 $\times$ 32& $\approx$ 0.0\\
				CIFAR-100 & 50K & - & 10K & 100 & 32 $\times$ 32 & $\approx$ 0.0\\
				\midrule
				\multicolumn{7}{c}{\multirow{1}{*}{Datasets with real world noisy annotation}} \\
				\midrule
				ANIMAL-10N & 50K & - & 5K & 10 & 64 $\times$ 64 & $\approx$ 8\\
				Clothing1M & 1M & 14K & 10K & 14 & 224 $\times$ 224 & $\approx$ 38.5\\
				Webvision& 66K & - &2.5K & 50 & 256 $\times$ 256& $\approx$ 20.0\\	
				\bottomrule
			\end{tabular}
		}
	\end{center}
	\caption{Description of the datasets used in the experiments.}
	\label{table:datasets}
\end{table*}

\subsection{Baselines}
We compare SELC to the following baselines from different categories. (1) CE directly uses the standard
cross-entropy loss to train the DNNs on noisy training data. (2) Forward \cite{patrini2017making} and Bootstrap \cite{reed2014training} belong to loss correction category. (3) GCE \cite{zhang2018generalized}, SCE \cite{wang2018iterative}, NCE+RCE \cite{ma2020normalized} and DMI \cite{xu2019l_dmi} belong to robust loss function category. (4) Joint Opt \cite{tanaka2018joint}, PENCIL \cite{yi2019probabilistic}, M-correction \cite{arazo2019unsupervised}, SELFIE \cite{song2019selfie}, SEAL \cite{chen2021beyond}, PLC \cite{zhang2020learning} and LRT \cite{zheng2020error} belong to label correction category. (4) Co-teaching \cite{han2018co}, Co-teaching+ \cite{yu2019does} Iterative-CV [60] belong to sample selection category. (4) NLNL \cite{kim2019nlnl}, DAC \cite{thulasidasan2019combating}, SELF \cite{nguyen2020self} and CRUST \cite{mirzasoleiman2020coresets} belong to noisy pruning category. (5) ELR \cite{liu2020early} and Nested \cite{chen2021boosting} belongs to regularization category. Note that we do not compare with some state-of-the-art methods like DivideMix \cite{li2019dividemix} and RoCL \cite{zhou2020robust} as baseline, because their proposed methods are aggregations of multiple techniques while this paper only focuses on one, therefore the comparison is not fair.

\subsection{Training Details on Class-conditional Label Noise.}
\textbf{Preprocessing.} We apply the standard data augmentation on CIFAR-10/100: horizontal random flip and 32 $\times$ 32 random crop after padding 4 pixels around images. The standard normalization with mean=(0.4914, 0.4822, 0.4465), std=(0.2023, 0.1994, 0.2010) is applied before feeding images to the network.

\begin{figure}[t]
	\begin{center}
		\includegraphics[width=0.8\linewidth]{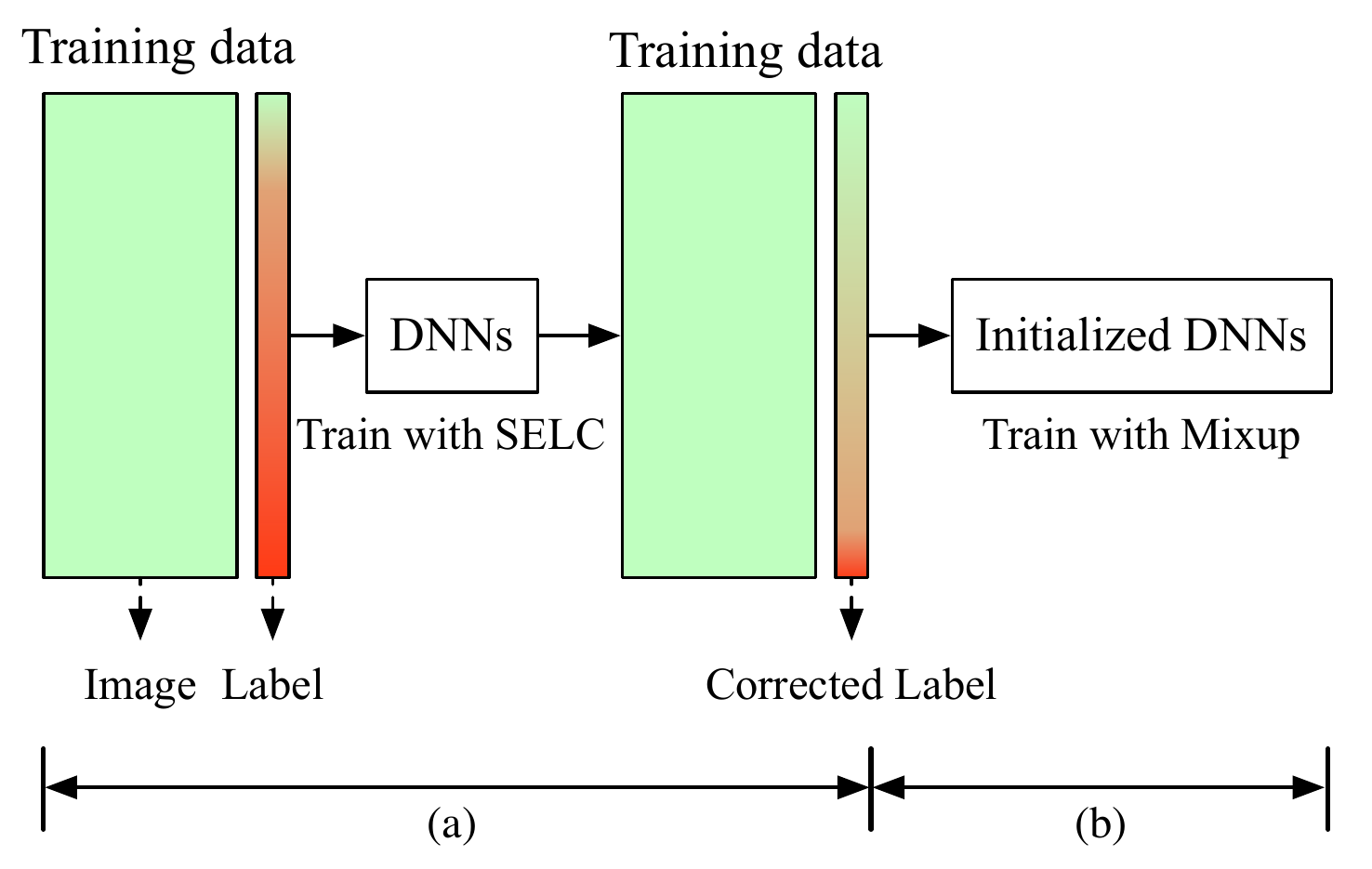}
	\end{center}
	\caption{(a): We train the DNNs with SELC and get the corrected label $\bm{t}$. (b): We train an initialized DNNs with mixup using the corrected labels from (a). }
	\label{fig:selc_mixup}
\end{figure}

We conduct the experiments with class-conditional label noise on CIFAR-10 and CIFAR-100 \cite{krizhevsky2009learning}. Given these two datasets are initially clean, we follow \cite{patrini2017making} to inject noise by label transition matrix $\mathbf{Q}$, where $\mathbf{Q}_{ij}=\Pr[\hat{y}=j\mid y=i]$ denotes the probability that noisy label $\hat{y}$ is flipped from clean label $y$. We evaluate SELC in two types of noise: \emph{symmetric} and \emph{asymmetric}. Symmetric noise is generated by randomly replacing the labels for a percentage of the training data with all possible labels. Asymmetric noise is designed to mimic the structure of real-world label noise, where the annotators are more likely to make mistakes only within very similar classes (e.g. \emph{deer} $\rightarrow$ \emph{horse} and \emph{cat} $\leftrightarrow$ \emph{dog}). We use the ResNet34 \cite{he2016deep} as backbone for both datasets, and train the model using SGD with a momentum of 0.9, a weight decay of 0.001, and a batch size of 128. The network is trained for 200 epochs. We set the initial learning rate as 0.02, and reduce it by a factor of 10 after 40 and 80 epochs. For parameter $\alpha$ in SELC, we fix $\alpha=0.9$. Note that we do not perform early stopping since we don’t assume the presence of clean validation data. All test accuracy are recorded from the last epoch of training. The illustration of SELC+ are shown in Figure \ref{fig:selc_mixup}.

\subsection{Training Details on Instance-dependent Label Noise.} 
For instance-dependent noise from SEAL, we use the same network architecture Wide ResNet28$\times$10 \cite{zagoruyko2016wide}. Models are trained for 150 epochs with a batch size of 128 and we report the test accuracy at the last epoch. We use SGD with a momentum of 0.9 and a weight decay of $5\times10^{-4}$. The learning rate is initialized as 0.1 and is divided by 5 after 60 and 120 epochs. We report the test accuracy at the last epoch.

As for PMD noise, we evaluate SELC on three types of label noise from the PMD noise family for consistency with PLC. We use the same network architecture PreAct ResNet34 \cite{he2016identity} as PLC. We use SGD with a momentum of 0.9 and a weight decay of $5\times10^{-4}$. For noise on CIFAR-10, models are trained for 150 epochs with a batch size of 128. The learning rate is initialized as 0.1 and is divided by 5 after 60 and 120 epochs. For noise on CIFAR-100, models are trained for 200 epochs with a batch size of 128. The learning rate is initialized as 0.02 and is divided by 10 after 40 and 80 epochs. 

\subsection{Training Details on Real-world Label Noise} 
\textbf{Preprocessing.} Following \cite{zhang2020learning}, for ANIMAL-10N, we apply normalization and regular data augmentation (i.e. horizontal flip) on the training sets. The standard normalization with mean=(0.485, 0.456, 0.406), std=(0.229, 0.224, 0.225) is applied before feeding images to the network. For Clothing1M and Webvision, we apply normalization and regular data augmentation (i.e. random crop and horizontal flip) on the training sets. The cropping size is consistent with existing works \cite{liu2020early}. Specifically, 224 × 224 for Clothing 1M (after resizing to 256 × 256), and 227 × 227 for Webvision. The standard normalization with mean=(0.6959, 0.6537, 0.6371), std=(0.3113, 0.3192, 0.3214) for Clothing1M and mean=(0.485, 0.456, 0.406), std=(0.229, 0.224, 0.225) for Webvision.

\noindent\textbf{ANIMAL-10N} contains 50,000 human-labeled online images for 10 animals with confusing appearance. The estimated label noise rate is 8\%. There are 50,000 training and 5,000 testing images. Following \cite{song2019selfie}, we use VGG-19 with batch normalization and train it using SGD with a weight decay 0.001, batch size 128 and initial learning rate 0.1, which is reduced by a factor of 5 after 50 and 75 epochs (100 epochs in total). We repeat the experiments with 3 random trials and report the mean value and standard deviation. All test accuracy are recorded from the last epoch of training. 

\noindent\textbf{Clothing1M} consists of 1 million training images collected from online shopping websites. The labels are generated by using the surrounding texts of the images that are provided by the sellers and thus contain many wrong labels. The estimated label noise rate is 38.5\%. It also contains 50k, 14k, and 10k of clean data for training, validation, and testing, respectively. Note that we do not use the 50k clean data in our training process. We use ResNet50 \cite{he2016deep} pretrained on ImageNet and train it using SGD with a momentum 0.9, weight decay 0.001, batch size 64 and initial learning rate 0.01, which is reduced by a factor of 10 after 10 and 20 epochs (30 epochs in total). For each epoch, we sample 2000 mini-batches from the training data ensuring that the classses of the noisy labels are balanced. We report the accuracy on the test set when the performance on the validation set is optimal. 

\noindent\textbf{Webvision} contains 2.4 million images crawled from the web using the 1,000 concepts in ImageNet ILSVRC12. The estimated label noise rate is 20\%. Following \cite{chen2019understanding}, we use the first 50 classes of Google image subset for training and test on the corresponding 50 classes of WebVision (approximate 66K) and ILSVRC-2012 validation set. We use InceptionResNetV2 \cite{szegedy2016rethinking} and train it using SGD with a momentum 0.9, weight decay 0.0005, batch size 32 and initial learning rate 0.01, which is reduced by a factor of 10 after 40 and 80 epochs (100 epochs in total). We report the accuracy on the Webvision validation set and ImageNet ILSVRC-2012 validation set.

\end{document}